\documentclass[conference]{IEEEtran}
\IEEEoverridecommandlockouts
\pdfoutput=1 

\usepackage{cite}
\usepackage{amsmath,amssymb,amsfonts}
\usepackage{algorithmic}
\usepackage{graphicx}
\usepackage{textcomp}
\usepackage{xcolor}

\usepackage{multicol}
\usepackage{multirow}

\usepackage{booktabs} 
\usepackage{adjustbox} 
\usepackage{tabularx} 
\usepackage{fancyhdr}
\def\BibTeX{{\rm B\kern-.05em{\sc i\kern-.025em b}\kern-.08em
    T\kern-.1667em\lower.7ex\hbox{E}\kern-.125emX}}
\begin{document}

\title{Conference Paper Title}

\title{Early Detection and Classification of Breast Cancer Using Deep Learning Techniques}


\author{
    \IEEEauthorblockN{Mst. Mumtahina Labonno, D.M. Asadujjaman, Md. Mahfujur Rahman, Abdullah Tamim,\\ Mst. Jannatul Ferdous, Rafi Muttaki Mahi}
    \IEEEauthorblockA{
        \textit{Dept. of Computer Science and Engineering, Varendra University, Rajshahi} \\
        mumtahina0614nurah@gmail.com, asadujjaman@vu.edu.bd, mahfujur@vu.edu.bd, \\
        abtamim131415@gmail.com, jannatul.cse10@gmail.com, raf@alphatechds.com
    }
}

\maketitle

\begin{abstract}
Breast cancer is one of the deadliest cancers causing about massive number of patients to die annually all over the world according to the WHO. It is a kind of cancer that develops when the tissues of the breast grow rapidly and unboundly. This fatality rate can be prevented if the cancer is detected before it gets malignant. Using automation for early-age detection of breast cancer, Artificial Intelligence and Machine Learning technologies can be implemented for the best outcome. In this study, we are using the Breast Cancer Image Classification dataset collected from the Kaggle depository, which comprises 9248 Breast Ultrasound Images and is classified into three categories: Benign, Malignant, and Normal which refers to non-cancerous, cancerous, and normal images.This research introduces three pretrained model featuring  custom classifiers that includes ResNet50, MobileNet, and VGG16, along with a custom CNN model utilizing the ReLU activation function.The models ResNet50, MobileNet, VGG16, and a custom CNN recorded accuracies of 98.41\%, 97.91\%, 98.19\%, and 92.94\% on the dataset, correspondingly, with ResNet50 achieving the highest accuracy of 98.41\%.This model, with its deep and powerful architecture, is particularly successful in detecting aberrant cells as well as cancerous or non-cancerous tumors. These accuracies show that the Machine Learning methods are more compatible for the classification and early detection of breast cancer.
\end{abstract}


\section{Introduction}
Carrying out one of the most deaths worldwide, cancer is responsible for nearly 10 million deaths and for ranking in terms of new cases, breast cancer was at the top having about 2.26 million cases in 2020\cite{b2}. In that year, about 685000 deaths were caused by the deadliest breast cancer. According to WHO, about 2.3 million women were diagnosed and 6.7 lac deaths worldwide in 2022 because of this cancer\cite{b1}. Both men and women can be affected but on average, 99\% of breast cancers occur in women and the rest of 0.5-1\% for men. So, women are at the highest risk for breast cancer \cite{b1}. In countries with an extremely high Human Development Index (HDI), about one in every 12 women will be diagnosed with breast cancer during their lifetime, and one in every 71 will die from the disease. In contrast, in nations with a low Human Development Index, the ratio changes to one in every 27 women receiving a breast cancer diagnosis, with one in every 48 succumbing to it. Detecting this cancer at an early stage can greatly improve recovery chances and lower death rates. When breast cancer is caught early, the five-year survival rate is 99\%. If the cancer progresses, the survival rate drops to 86\%, whereas late-stage discovery leads to only a 31\% survival rate.\cite{b3}.
Breast cancer is defined as abnormalities in breast tissues caused by rapid and unregulated tissue growth, which leads to the formation of tumors, which can be benign or malignant. The malignant tumors can spread throughout the lymphatic or circulatory system, resulting in breast cancer. For detecting if a cell is benign or malignant, it is necessary to analyze that tissue through an image to identify its characteristics, such as its size, irregular growth, shape, or textures. Due to this, in this study, 9248  Breast Ultrasound Images from the Breast Cancer Image Classification dataset were collected from the Kaggle depository, which is Benign, Malignant, and Normal. A normal image means healthy breast tissues, a benign image represents non-cancerous breast tissues, and a malignant image for diseased or cancerous tissues. ResNet50, a type of convolutional neural network, was trained on those images to achieve optimal accuracy in detecting breast cancer. This deep learning model is designed to excel at analyzing the fine aspects of medical photos. This model consists of 50 layers including convolutional, pooling, fully connected, and batch normalization layers with ReLu activation functions for extracting figures resizing images, and classification training. It is a pre-trained model that is very flexible to medical images. This approach is effective and provides greater scalability due to its robust and extensive architecture. In this research, the accuracy achieved is 98.41\% with ResNet50, while MobileNet, VGG16, and a custom CNN registered accuracies of 97.91\%, 98.19\%, and 92.94\%, respectively. This indicates that employing deep learning techniques like ResNet50 for the early detection of breast cancer is not only highly efficient but also crucial for saving lives.

\section{Literature Review}
Due to being a well-known deadliest cancer all over the world, several research projects and experiments have been conducted for the early detection of breast cancer.For those researches, there used different models from machine learning and deep learning methods like Logistic Regression (LR), Random Forest, K-Nearest Neighbors (KNN), Support Vector Machine (SVM), Artificial Neural Networks (ANN), Convolutional Neural Networks (CNN) etc for analyzing structured and unstructured complex data with best adaptability feature. Machine learning models manually extract features from data whereas deep learning models automatically extract relevant features from images. Different datasets from UCI Machine Learning Repository and, Kaggle Depository are used in those research fields. Non-image data includes text, categorical, and numerical data, while mammography, MRI, histopathology, and ultrasound images are categorized as image datasets.

Motivated by the advanced deep learning architectures GoogleNet and residual blocks, the study of Salman Zakareya et al. introduces a novel deep model aimed at breast cancer classification, achieving an accuracy rate of 93\% for ultrasound images and 95\% for breast histopathology images\cite{b4}. 

The ResNet50 CNN model on the INbreast dataset with mammogram images was conducted by Hameedur Rahman et al. and published in 2023 with an accuracy of 93\% \cite{b5}.

Yanhui Guo et al. applied the model called FRPC, which stands for Fuzzy Relative Position Coding Transformer, using a benchmark dataset for breast ultrasound images and achieved an accuracy of 90.52\%\cite{b6}.

Multiple pre-trained deep learning models named Xception, InceptionV3, VGG16, MobileNet, \& ResNet50 scored an F1 score accuracy of 97.54\%, 95.33\%, 98.14\%, 97.67\%, \& 93.98\%, respectively, while BCCNN, which was the proposed model, achieved an accuracy of 98.30\%  and 98.28\% for recall in  the early detection of breast cancer was received by Basem S Abunasser et al. with the breast cancer history pathological image classification dataset, which is also known as the breakHis dataset from Kaggle Respiratory \cite{b7}. 

From the paper submitted by Meshrif Alruily et al. developed a vision transformer (ViT) model utilizing a breast ultrasound images dataset having 780 images for benign and malignant breast tissues, achieving an accuracy of 94.49\%. This model is employed to manage intricate capital relationships effectively, with the potential for enhanced accuracy \cite{b8}.

With a motive to identify which models are based for early detection of breast cancer a Research named Competitive Analysis of deep learning architectures for breast cancer diagnosis using breaKHis Dataset conducted by Irem Sayin et al. compared the performance of five deep learning models for cancer VGG, ResNet, Xception, Inception, and InceptionResNet, and the top accuracy was obtained by the Xception model with an accuracy of 89\% where Inception and InceptionResNet both has the accuracy of 87\% \cite{b9}.

This study employs a deep learning method using the ResNet50 architecture along with three more models, focusing on the precise and efficient identification and classification of breast cancer in its initial stages

\section{Methodology}

\subsection{Dataset and Experiment}

\begin{table}[htbp]
\caption{Details of Dataset}
\centering
\begin{tabular}{c c c c c }
\hline
\textbf{Class} & \textbf{ultrasound/ Class} & \multicolumn{3}{ c }{\textbf{Dataset splitting}} \\
\cline{3-5}
&  & \textbf{Train} & \textbf{Validation} & \textbf{Test} \\
\hline
Benign     & 4711 & 3297 & 707 & 707 \\
Malignant         & 4271 & 2989 & 641 & 641 \\
Normal & 266 & 186 & 40 & 40 \\
\hline
\end{tabular}
\label{tab1}
\end{table}

\begin{figure*}[htbp]
\centerline{
\includegraphics[width=0.85\textwidth, height=0.15\textheight]{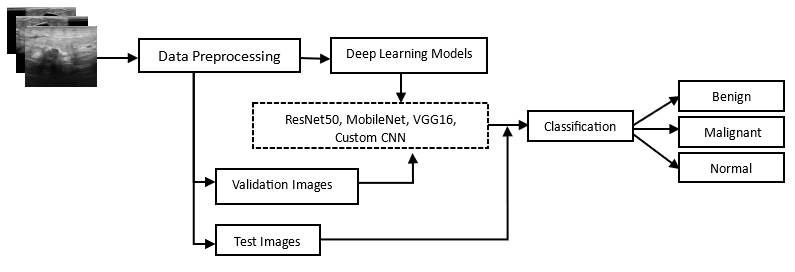}}
\caption{Block diagram of the multi-class classification of proposed framework}
\label{blockdiagram}
\end{figure*}

In this research, the dataset originates from the Kaggle repository, the Breast Cancer Image Classification dataset. This dataset comprises 9,248 ultrasound images that are labeled as benign, malignant, and normal, with counts of 4,711, 4,271, and 266 images respectively. This comprehensive collection of data is extremely valuable for AI-powered diagnostic solutions, as it offers labeled images that support the early identification of breast cancer.

\subsection{Data Pre-Processing}
All images in the dataset were resized to dimensions of $224\times224$ pixels, after which they were stored in an array for additional processing that involved reading from their file paths.
The dataset was partitioned into training, validation, and test sets, with 70\% allocated for training and the remaining 30\% evenly distributed between validation and test datasets to evaluate model performance. 

\subsection{Training Models}
\subsubsection{Deep Transfer Learning Model}
We used pretrained models ResNet50, MobileNet, and VGG16 each with a custom classifier head to classify data based on specific goals.

\begin{itemize}
\item ResNet50: A 50-layer deep network using residual connections to capture hierarchical features, making it effective for large datasets and deep networks.

\item MobileNet: Designed for mobile and embedded devices, it balances speed and accuracy by using depthwise and pointwise convolutions to reduce computational cost, suitable for real-time applications.

\item VGG16: A 16-layer CNN known for its simplicity, using 3x3 convolutional layers and max pooling, making it effective for visual tasks but computationally demanding.
\end{itemize}

 The classification head consists of a global average pooling layer followed by a fully connected dense layer with 1024 units and ReLU activation, and a final softmax layer for multi-class classification.\\

\subsubsection{Custom CNN}
\begin{figure}[htbp]
\centerline{\includegraphics[width=0.5\textwidth, height=0.2\textheight]{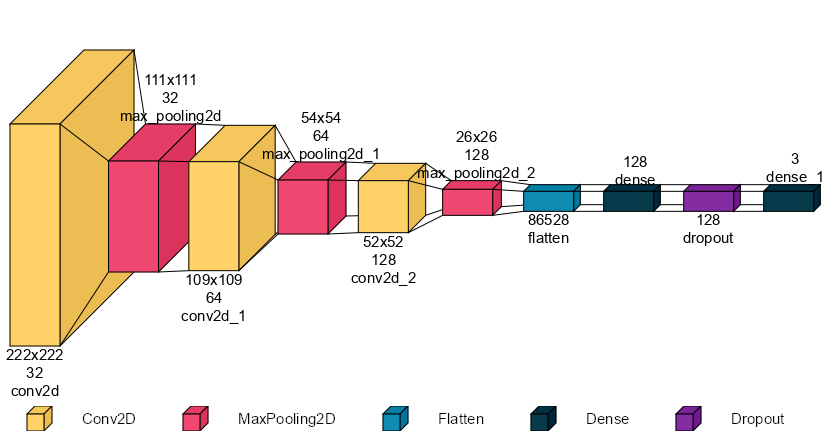}}
\caption{Custom CNN Architecture}
\label{customcnn}
\end{figure}

A custom convolutional neural network refers to a CNN architecture that is tailored to meet the unique needs of a specific dataset. In this study, a Custom CNN model was employed with multiple layers configured in various ways, followed by a ReLU activation function that injects nonlinearity into the network shown in Figure \ref{customcnn}. ReLU helps address the vanishing gradient issue during the training of deep learning models, allowing for the learning of more intricate relationships within the data.

\subsection{Hyperparameters}
Hyperparameters control the learning process of models in machine or deep learning and are set prior to the training phase. These parameters can be tailored for specific tasks and play a crucial role in managing learning processes, minimizing overfitting and underfitting, and enhancing overall performance. Table \ref{tab2} presents the hyperparameters associated models.

\begin{table}[htbp]
\caption{Models hyperparameters for input and classification stage}
\centering
\begin{tabular}{c c c}
\hline
\textbf{Parameters}  & \multicolumn{2}{c}{\textbf{Approch}} \\
\cline{2-3}
  & \textbf{Pretrained Model} & \textbf{Custom CNN} \\
\hline
Input shape & 224 $ \times $ 224 & 224 $ \times $ 224 \\
No. of epochs & 10 & 10 \\
Batch Sizes & 32 & 32 \\
Activation Function & Softmax & Softmax \\
Learning rate & 0.0001 & 0.0001 \\
\hline
\end{tabular}
\label{tab2}
\end{table}

\subsection{Evaluation Criteria}
Accuracy: Accuracy is assessed by the proportion of accurate predictions (true positives and true negatives) produced by a model. This metric is simple and extremely useful when the class distribution is balanced. 
\begin{equation}
\text{Accuracy} = \frac{TP + TN}{TP + TN + FP + FN}
\end{equation}

F1 Score: The F1 score  is determined by the harmonic mean of precision and recall.The score can range from zero to one, with a score of one reflecting perfect precision and recall.
\begin{equation}
\text{F1 Score} = 2 \times \frac{\text{Precision} \times \text{Recall}}{\text{Precision} + \text{Recall}}
\end{equation}
Precision: Precision, defined as the ratio of true positive results to all instances that are predicted as positive, serves as a measure to assess the accuracy of positive predictionscalculated by dividing the total count of positive predictions (TP + FP) by the count of true positives.
\begin{equation}
\text{Precision} = \frac{TP}{TP + FP}
\end{equation}
Recall: Recall guarantees a majority of true positives are detected. Recall is calculated by dividing the number of true positives (TP) by the total number of actual positives (TP + FN).
\begin{equation}
\text{Recall} = \frac{TP}{TP + FN}
\end{equation}
AUC-ROC: AUC (Area Under the Curve) and ROC (Receiver Operating Characteristic) are metrics used to evaluate classification models by assessing their ability to distinguish between classes across all thresholds. A higher AUC value indicates better performance, with a value closer to 1.0 representing a strong model and 0.5 indicating random guessing.

\section{Result And Discussion}
In this study, we assessed the effectiveness of three deep learning models ResNet50, MobileNet, VGG16, and a custom CNN model  with the ReLU activation function. The ResNet50 model demonstrated excellent performance on the dataset, achieving high accuracy of 98.41\% shown in Table \ref{tab3}.

\begin{table}[htbp]
\caption{Model comparison across studies}
\begin{center}
\begin{tabular}{ c c c c} 
\hline
\textbf{Study no.} & \textbf{Models used} & \textbf{Accuracy} & \textbf{F1 Score} \\
\hline
\textbf{\cite{b5}} & ResNet50 & 0.93 & 0.9303\\
\hline
\textbf{\cite{b6}} & FRPC & 0.9052 & 0.9052\\
 \hline

\textbf{\cite{b7}} & BCCNN & 0.9830 & 0.9828\\
\hline
\multirow{3}{*}{\textbf{\cite{b9}}} 
& Xception & 0.89 & 0.90\\
& Inception & 0.87 & 0.87\\
& InceptionResNet & 0.87 & 0.86\\
\hline
\multirow{4}{*}{\textbf{This study}} 
& ResNet50 & 0.9841 & 0.96\\
& MobileNet & 0.9791 & 0.97\\
& VGG16 & 0.9819 & 0.963\\
& Custom CNN & 0.9294 & 0.91\\
\hline
\end{tabular}
\label{tab4}
\end{center}
\end{table}

\begin{figure}[htbp]
\centerline{\includegraphics[width=0.4\textwidth, height=0.3\textheight]{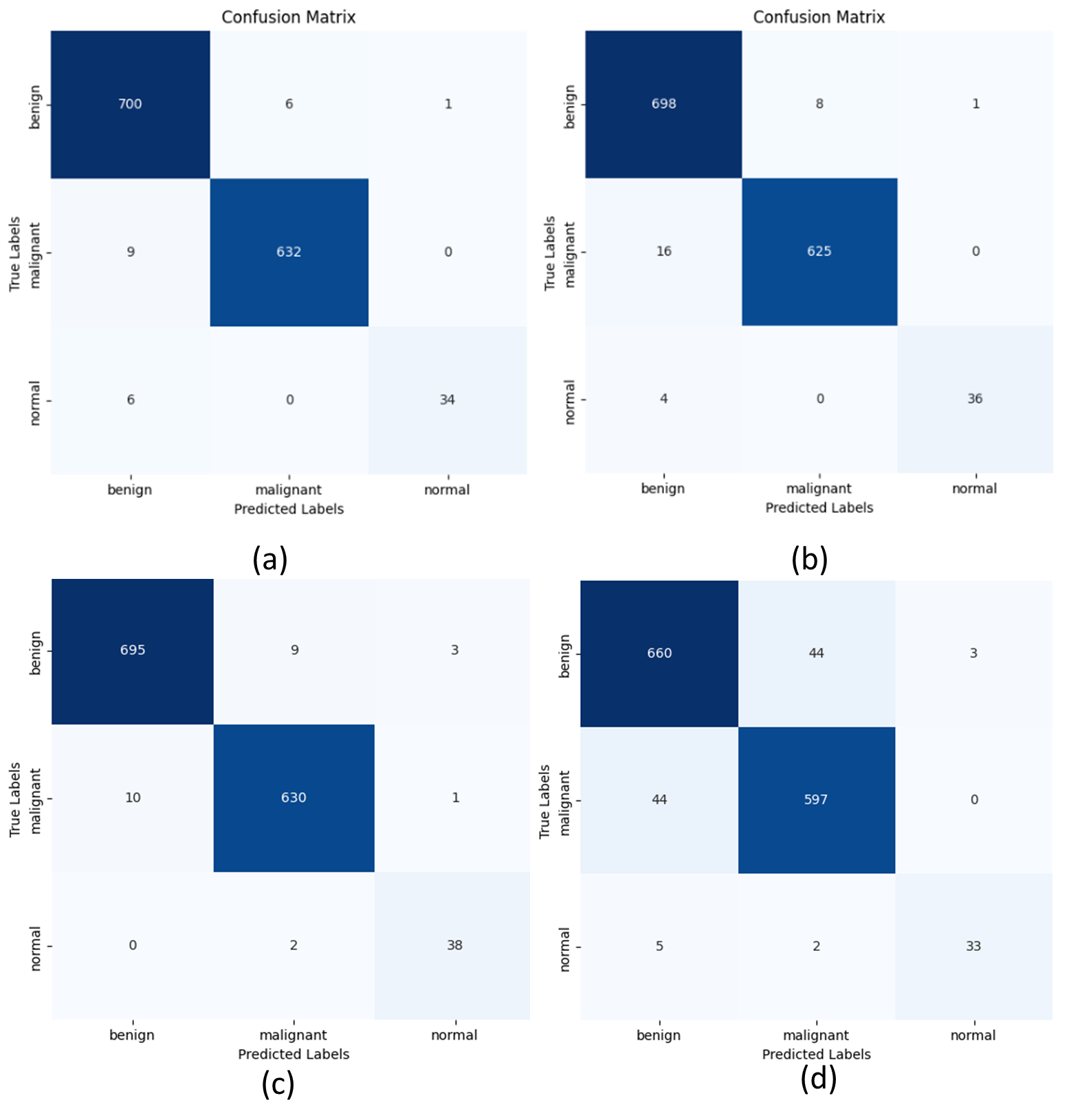}}
\caption{Confusion Matrix Visualization for (a)ResNet50, (b)MobileNet, (c)VGG16, and (d) custom CNN model }
\label{confusionmatrix}
\end{figure}
The confusion matrix in Figure \ref{confusionmatrix} evaluates the classification technique's performance on an imbalanced dataset. Unlike accuracy, which can be misleading, it offers deeper insights into the model's ability to handle both majority and minority classes effectively. It shows that the model avoids bias toward the majority class, addressing potential issues like overfitting or under-representation.

\begin{figure}[htbp]
\centerline{\includegraphics[width=0.4\textwidth, height=0.3\textheight]{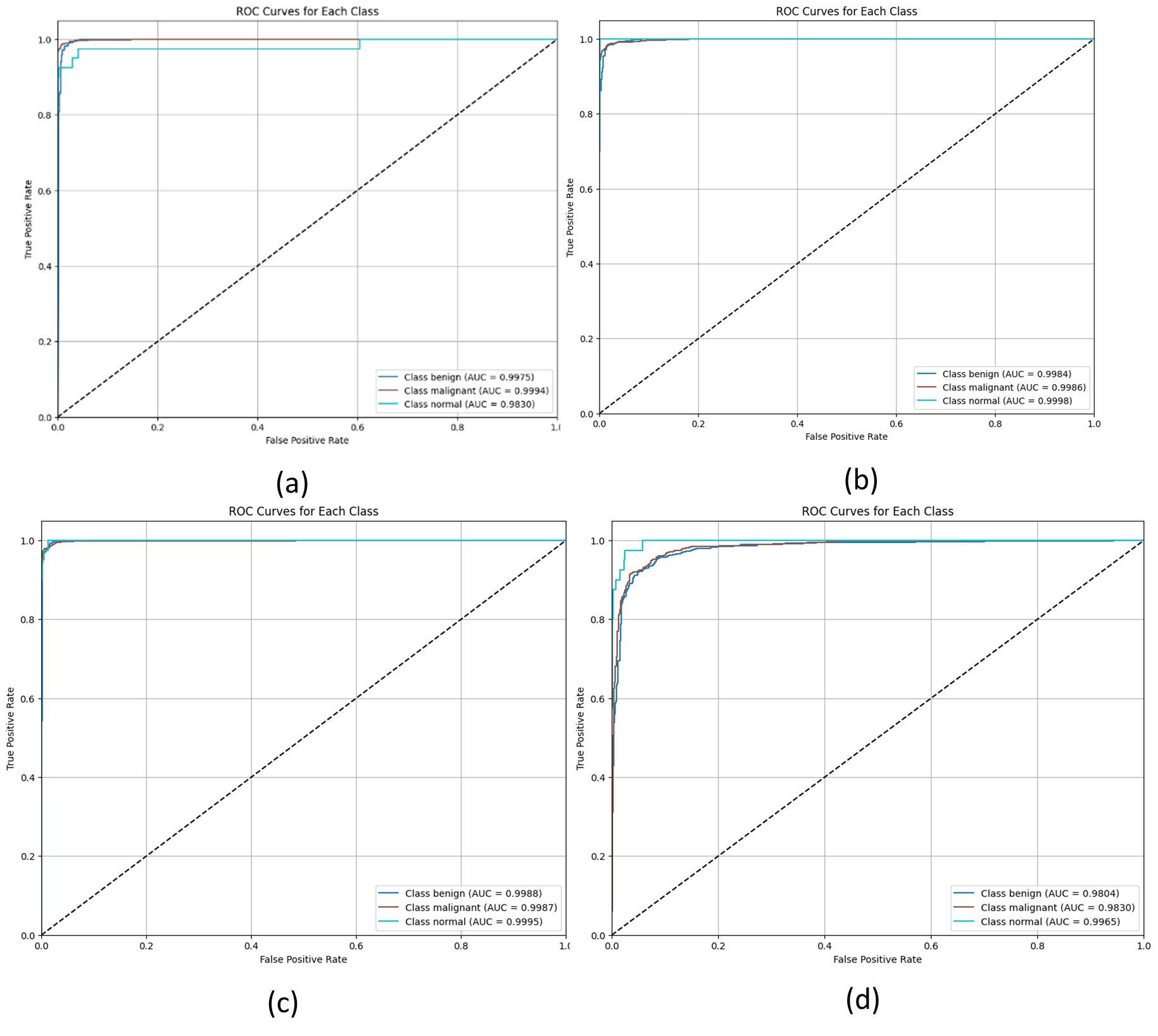}}
\caption{ROC curve of ROC curves for (a)ResNet50, (b)MobileNet, (c)VGG16, and (d) custom CNN model }
\label{roccurve}
\end{figure}

Figure \ref{roccurve} illustrates the ROC curves for ResNet50, MobileNet, VGG16, and a custom CNN model. The analysis indicates   that VGG16 achieves the highest AUC of 0.9990, followed by MobileNet 0.9989, ResNet50 0.9933, and the custom CNN model with the lowest AUC of 0.9867.
\begin{table}[htbp]
\centering
\caption{Model Performance}
\adjustbox{max width=\textwidth}{
\begin{tabular}{ c c cc c }
\hline
\multirow{2}{*}{\textbf{Model}} & \multirow{2}{*}{\textbf{Class}} & 
\multicolumn{2}{c }{\textbf{Dataset}}  & \multirow{2}{*}{\textbf{Accuracy}}\\
\cline{3-4}
 &  & \textbf{Precision} & \textbf{Recall} \\
\hline
\multirow{3}{*}{ResNet50} 
    & Benign    & 0.98 & 0.99 & \multirow{3}{*}{0.9841}  \\
    & Malignant & 0.99 & 0.99 &                      \\
    & Normal    & 0.97 & 0.85 &                      \\
\hline
\multirow{3}{*}{MobileNet} 
    & Benign    & 0.97 & 0.99 & \multirow{3}{*}{0.9791}  \\
    & Malignant & 0.99 & 0.98 &                      \\
    & Normal    & 0.97 & 0.90 &                      \\
\hline

\multirow{3}{*}{VGG16} 
    & Benign    & 0.99 & 0.98 & \multirow{3}{*}{0.9819}  \\
    & Malignant & 0.98 & 0.98 &                      \\
    & Normal    & 0.90 & 0.95 &                      \\
\hline
\multirow{3}{*}{Custom CNN} 
    & Benign    & 0.93 & 0.93 & \multirow{3}{*}{0.9294}  \\
    & Malignant & 0.93 & 0.93 &                      \\
    & Normal    & 0.92 & 0.82 &                      \\
\hline
\end{tabular}
}
\label{tab3}
\end{table}

Table \ref{tab3} highlights the performance of four models for early breast cancer detection. ResNet50 achieved the highest accuracy 98.41\% but struggled with normal case recall 0.85. MobileNet 97.91\% accuracy improved recall for normal cases 0.90. VGG16 showed balanced performance 98.19\% accuracy, 0.98 precision and recall for benign and malignant cases. The custom CNN had the lowest accuracy 92.94\% with consistent precision and recall for benign and malignant cases but lower recall 0.82 for normal cases.

Table \ref{tab4} indicates that the study compares the performance of various existing models, including ResNet50, MobileNet, VGG16, and Custom CNN, achieving high accuracy and F1 scores, with ResNet50 performing the best in both metrics, followed by MobileNet and VGG16.

\section{Conclusion}
Breast cancer causes a significant number of deaths each year and brings about various difficulties for patients upon receiving a diagnosis, impacting their physical and mental health. If the disease is detected before it becomes malignant, the suffering experienced can be alleviated. As shown in the study mentioned earlier, deep learning techniques can be utilized quite effectively in this area. While we recognize that no technology can replace a doctor’s skill and experience, our intention is not to supplant physicians but rather to use deep learning methods and models in the analysis of the dataset, aiding in the early detection or identification of the disease, which can ultimately lessen the burden on patients. This approach can also support doctors in making more accurate decisions, reducing the risk of misdiagnoses and preventing the premature prescription of medications or treatments.

\end{document}